\documentclass[runningheads,dvipsnames]{llncs}

\newcommand{\visBasePath}[0]{images/}

\usepackage[utf8]{inputenc} 
\usepackage[english]{babel}
\usepackage{graphicx} 
\usepackage{xspace} 
\usepackage{xcolor}
\usepackage{ifthen}
\usepackage{booktabs}
\usepackage{multirow}
\usepackage{multicol}
\usepackage{wrapfig}
\usepackage{url}

\newboolean{vc_is_included}
\InputIfFileExists{vc.tex}{
	\setboolean{vc_is_included}{true}
}{
	\setboolean{vc_is_included}{false}
	
	\gdef\GITAbrHash{}

	\gdef\VCDateTEX{}

}

\newboolean{requirement_jmlr}
\ifthenelse{\isundefined{\makeJMLR}}{
	\setboolean{requirement_jmlr}{false}}{
	\setboolean{requirement_jmlr}{true}}

\newboolean{requirement_nosubfig}
\ifthenelse{\isundefined{\makeWithoutSubfig}}{
	\setboolean{requirement_nosubfig}{false}}{
	\setboolean{requirement_nosubfig}{true}}

\ifthenelse{\boolean{requirement_nosubfig}}{
}{
    \ifthenelse{\boolean{requirement_jmlr}}{
    }{
        \usepackage{subfig}
    }
}

\newboolean{article_extended}
\ifthenelse{\isundefined{\makeExtended}}{
	\setboolean{article_extended}{false}}{
	\setboolean{article_extended}{true}}

\newboolean{requirement_blinded}
\ifthenelse{\isundefined{\makeBlinded}}{
	\setboolean{requirement_blinded}{false}}{
	\setboolean{requirement_blinded}{true}}

\newboolean{requirement_twocolumn}
\ifthenelse{\isundefined{\makeRequirementTwocols}}{
	\setboolean{requirement_twocolumn}{false}}{
	\setboolean{requirement_twocolumn}{true}}

\ifthenelse{\isundefined{\makeWithoutHref}}{
    \usepackage[bookmarks=false]{hyperref}
    \hypersetup{colorlinks,
      linkcolor=blue,
      citecolor=blue,
      urlcolor=blue}
}{}

\newcommand{\articleTitle}[0]{Efficient NAS with FaDE on Hierarchical Spaces}

\newcommand{\articleSubtitle}[0]{}

\newcommand{\articleAuthorSpringer}[0]{Simon Neumeyer \and Julian Stier \orcidID{0000-0001-5710-9240} \and Michael Granitzer \orcidID{0000-0003-3566-5507}}
\newcommand{\articleAuthorRunning}[0]{Neumeyer et al.}
\newcommand{\articleEmail}[0]{neumeyer.simon@gmx.de\protect\\julian.stier@uni-passau.de}
\newcommand{\articleInstituteSpringer}[0]{
University of Passau\\
\email{\articleEmail}
\ifthenelse{\boolean{vc_is_included}}{{\tiny\protect\\\tiny\VCDateTEX~$\sim$~\GITAbrHash}\vspace{-1em}}{~}
}

\DeclareSymbolFont{extraup}{U}{zavm}{m}{n}
\DeclareMathSymbol{\varheart}{\mathalpha}{extraup}{86}
\DeclareMathSymbol{\vardiamond}{\mathalpha}{extraup}{87}

\definecolor{Green}{rgb}{0.01, 0.75, 0.24}
\definecolor{Red}{rgb}{0.76, 0.23, 0.13}
\definecolor{Gray}{rgb}{0.41, 0.41, 0.41}

\newcommand{\nameFade}{\emph{FaDE}\xspace}

\newcommand{\jupyterLink}[1]{\href{#1}{\frame{\includegraphics[height=1em]{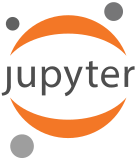}}}}
\newcommand{\urlNotebook}[0]{https://github.com/SimonNeumeyer/FaDE}
\newcommand{\urlNotebookA}[0]{\urlNotebook/Code/Theses_v1/jupyter/experimental.ipynb}
\newcommand{\urlNotebookB}[0]{\urlNotebook/Code/Theses_v1/jupyter/visualize_results.ipynb}

\usepackage{amsmath,amsfonts,bm}

\def\eqref#1{equation~\ref{#1}}

\def\1{\bm{1}}

\DeclareMathAlphabet{\mathsfit}{\encodingdefault}{\sfdefault}{m}{sl}
\SetMathAlphabet{\mathsfit}{bold}{\encodingdefault}{\sfdefault}{bx}{n}

\newcommand{\E}{\mathbb{E}}

\DeclareMathOperator*{\argmin}{arg\,min}

\makeatletter
\newcommand*{\inlineequation}[2][]{%
  \begingroup
    \refstepcounter{equation}%
    \ifx\\#1\\%
    \else
      \label{#1}%
    \fi
    \relpenalty=10000 %
    \binoppenalty=10000 %
    \ensuremath{%
      #2%
    }%
    ~\@eqnnum
  \endgroup
}
\makeatother

\usepackage[T1]{fontenc}
\begin{document}
\title{\articleTitle}
\titlerunning{\articleSubtitle}
%
\author{\articleAuthorSpringer}
\authorrunning{\articleAuthorRunning}
%
\institute{\articleInstituteSpringer}

\maketitle              
\begin{abstract}
Neural architecture search (NAS) is a challenging problem.
Hierarchical search spaces allow for cheap evaluations of neural network sub modules to serve as surrogate for architecture evaluations.
Yet, sometimes the hierarchy is too restrictive or the surrogate fails to generalize.  
We present FaDE which uses differentiable architecture search to obtain relative performance predictions on finite regions of a hierarchical NAS space.
The relative nature of these ranks calls for a memory-less, batch-wise outer search algorithm for which we use an evolutionary algorithm with pseudo-gradient descent.
FaDE is especially suited on deep hierarchical, respectively multi-cell search spaces, which it can explore by linear instead of exponential cost and therefore eliminates the need for a proxy search space.

Our experiments show that firstly, FaDE-ranks on finite regions of the search space correlate with corresponding architecture performances and secondly, the ranks can empower a pseudo-gradient evolutionary search on the complete neural architecture search space.

    \keywords{darts \and hierarchical neural architecture search \and automl \and differentiable structure optimization}
\end{abstract}

%
%
\section{Introduction}
\label{sec:introduction}
Automatically finding structures of deep neural architectures is an active research field.
The exponentially growing space of directed acyclic graphs (DAGs), their complex geometric structure and the expensive architecture performance evaluation make \textbf{neural architecture searches (NAS)} a challenging problem.
Methods such as evolutionary and genetic algorithms, bayesian searches and differentiable architecture searches compete for the most promising automatic approaches to conduct neural architecture searches \cite{elsken2019neural}.

Differentiable architecture search (DARTS) is a successful and popular method to relax the search space into a differentiable hyper-architecture.
This relaxation allows to use differentiable search methods to learn both model weights and architectural parameters to evaluate sub-paths of a hyper-architecture \cite{liu2018darts}.
While the combined and weight-shared hyper-architecture allows for a very fast training of few GPU days, the search space is limited to subspaces of the defined hyper-architecture.
Evolutionary searches, in contrast, are way more dynamic in the way they restrict the search space.
Without any weight-sharing tweaks, this usually comes with a significant higher computational time.


We present a \emph{FAst Darts Estimator on hierarchical search spaces} (\nameFade) that aims at optimizing chained like hierarchical architectures while not resorting to a proxy domain.
By iteratively fixing a finite set of sub-module architectures per neural sub-module and using DARTS for training, architecture ranks $\alpha$ are estimated based on the relative performance within an iteration.
As the estimations are of relative nature, we require a state-less, batch-wise optimization algorithm to determine from those estimations a new finite set of cell architectures per cell.
To this end, we apply an evolutionary approach which incorporates a pseudo-gradient descent for candidate generation.
\nameFade runs one independent optimization algorithm per cell which allows it to optimize additional depth with linear instead of exponential cost.

Our contributions contain the first usage of differentiably obtained ranks for neural architecture search in an open-ended search space.
The usage is justified with a correlation analysis.
We provide code and data of our experiments for reproducibility in a \href{\urlNotebook}{github repository}.
\nameFade might be generalized to further types of hierarchical search spaces and could also be employed with other state-less, batch-wise search strategies in open-ended search spaces.

%
%
\section{Fast DARTS Estimator} 
\label{sec:hierarchical-darts}
We construct chained hierarchical search spaces for neural architectures and use their (relative) estimated performance as architectural ranks.
The obtained \nameFade-ranks guide a search on the complete (but open) search space.
The structure of finite regions is not arbitrary, but bounded to the set of architectures contained in a hyper-architecture.
We use DARTS to train such a hyper-architecture, \nameFade to predict the corresponding region of the search space from a trained hyper-architecture, and a mapping of the search space into Euclidean space together with a pseudo gradient descent to guide the exploration of the search space.

\paragraph{On Chained Hierarchical NAS Spaces}
Motivated by repeated motifs in hand-crafted architectures, \cite{zoph2018} introduce hierarchy to NAS spaces by considering an architecture to be constructed from several structurally identical sub-modules, so called \textit{cells}.
By \emph{identical} the same architecture, yet each cell with its proper weights, is meant.
A cell typically consists of several convolutional layers, each with a variable operation type and with variable connections between layers.
Optimizing the \textit{cell architecture} is often equivalent to finding the type of convolutional operation for a fixed number of layers and determining which layers are being connected.
The \textit{macro-architecture} determines how multiple, often identical, cells are being stacked to one complete architecture.
Quite often, the stacking is done in e.g. a simple chain structure \cite{zoph2018}.
Many search strategies on hierarchical search spaces fix the macro-architecture and solely optimize the architecture of a single cell which is often done on a shallower \textit{proxy domain}.
For example, in image classification, a proxy domain might consist in switching from \textit{CIFAR-100} to \textit{CIFAR-10}.
Even though such search strategies can achieve very good results, they are not always successful.
On the one hand, the performance of a single cell might not generalize to the performance of an overall neural network architecture consisting of multiple structural copies of that cell.

\begin{figure*}[tbh]
    \vspace{-1em}
    \centering
    \includegraphics[width=0.9\linewidth]{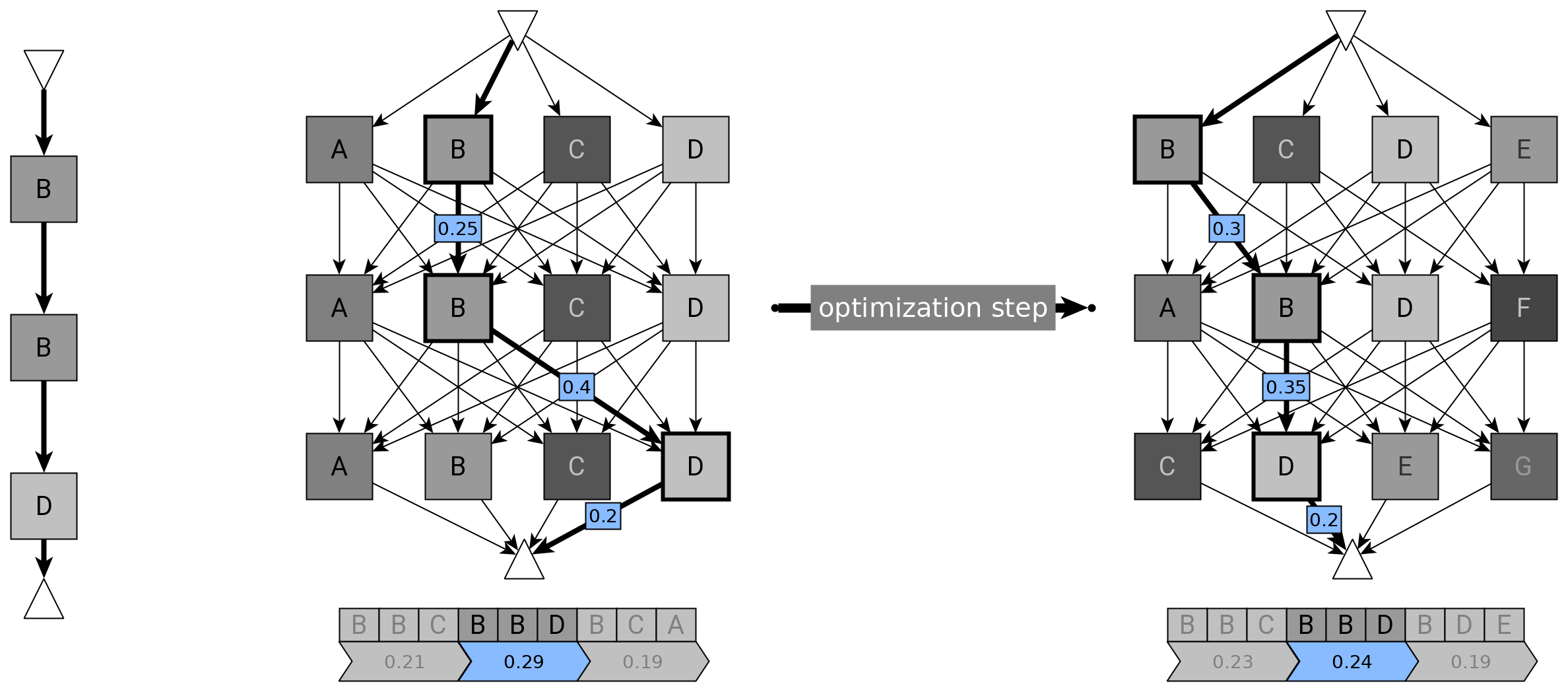}
    \label{fig:search_space}
    \caption{
    \small
    (left) Discrete architecture \textit{BBD} $\in \mathcal{S}^3$ featuring cell architectures $B,D\in\mathcal{S}$.
    (middle) \textit{BBD} contained in a hyper-architecture $H\in\mathcal{H}_{3,4}(\mathcal{S})$ that allows for several cell architectures per row. Obtaining relative \nameFade-ranks on trained hyper-architecture: factorizing architecture parameters along the corresponding path of the hyper-architecture.
    (right) Each step in the outer NAS optimization discovers new cell architectures per row.
    }
    \label{fig:search_space}
    \vspace{-3em}
\end{figure*}

\subsection{DARTS}
\label{darts_basics}
DARTS relaxes discrete architecture decisions within a differentiable hyper-architecture.
Liu et al. \cite{liu2018darts} consider a neural network module consisting of a set of edges $E$, that densely connects several ordered vertices, and a finite set of convolutional operations $O$.
The goal is to find the top $k\in \mathbb{N}$ incoming edges and respective operation per vertex.
To this end, \cite{liu2018darts} dedicate one architecture parameter $\alpha_{e,o}$ per edge $e\in E$ and operation $o\in O$, and calculate the output for any edge $e\in E$ as
\begin{equation}
\label{darts_agg_softmax}
    Softmax(\alpha_e)^\top o(.), o\in O
\end{equation}
Finding the best architecture parameters $\alpha$ aims at solving the bi-level optimization problem
\begin{equation}
    \label{bi-level}
    \argmin_{\alpha}L_{val}(\alpha, \argmin_{\omega}L_{train}(\alpha, \omega))
\end{equation}
with $L_{val}$ and $L_{train}$ denoting the neural network loss on a dedicated set of training samples each.
\cite{liu2018darts} propose a \textit{second-order} and a simplified \textit{first-order} objective for solving equation \ref{bi-level} via gradient descent.
The latter can be implemented by alternatingly fixing $\alpha$ and training $\omega$ on its respective split of training samples and vice verca.
After training, they derive an optimal discrete model by keeping only those operations with large architecture parameters.
A driving factor of \cite{liu2018darts} becoming the baseline work of differentiable architecture search consists in their bi-level optimization algorithm and in their choice of search space.
Liu et al. work on a cell from \cite{zoph2018} and further adopt their proxy concept, consisting of a smaller dataset and a smaller chain of cells during training.

\textbf{Gumbel-Softmax}
\cite{cai2018} criticize the Softmax in equation \ref{darts_agg_softmax} learning a well performing combination of architectures to which no single architecture will generalize once selected after training.
Instead, they propose a sampling mechanism in order to only activate the connections of a single architecture during any forward pass, while still applying back-propagation to the complete hyper-architecture.
To this end, instead of Softmax they apply \inlineequation[gumbel_softmax]{
    Gumbel\text{-}Softmax_\tau(x):=\allowbreak Softmax_\tau(\allowbreak x + G^n)
} to $x\in\mathbb{R}^n$ where $\tau>0$ is a temperature parameter for Softmax and $G^n$ is an i.i.d. vector of the \textit{standard Gumbel} distributed random variable $G$.
Gumbel-Softmax adds stochastic noise to a Softmax transformation while supporting differentiability.
For their forward pass they use an additional one-hot encoding on equation \ref{gumbel_softmax}.
\cite{jang2016categorical} show that it holds for any $\tau > 0, x\in \mathbb{R}^n$: 
$\E[Onehot \circ Gumbel\text{-}Softmax_\tau(x)] = Softmax(x)$

\subsection{Training a Chained Hierarchical Architecture using DARTS}
\label{method_applying_darts}
For an abstract space of cell architectures $\mathcal{S}$ and a depth $d\in\mathbb{N}$, we build the chained search space $\mathcal{S}^d$, see \autoref{fig:search_space}.
Given a window size $w\in\mathbb{N}$, we consider the corresponding space of hyper-architectures as $\mathcal{H}:=\mathcal{H}_{d,w}(\mathcal{S}):=\mathcal{S}^{d\times w}$, see \autoref{fig:search_space} (middle).
We use a matrix notation for hyper-architectures for convenience.
Note, that any hyper-architecture $H\in \mathcal{S}^{d\times w}$ can be identified with a subset of the search space by considering $\times_{i\leq d}\{H_{ij} ~|~ j\leq w\}\subset\mathcal{S}^d$, the cross product modelling the combinatorics of chaining cells along the depth of the search space.
Hence, a row of $H\in\mathcal{S}^{d\times w}$ represents the pool of cells that $H$ features at the corresponding depth.

Using differentiable architecture search (DARTS) \cite{liu2018darts}, we endow a hyper-architecture $H\in\mathcal{H}$ with architecture parameters $\alpha\in[0,1]^{d\times w}$, such that after training $H$ with a bi-level optimization algorithm, architecture parameter $\alpha_{ij}$ reflects the performance of cell $H_{ij}$ in the context of exclusive competition within each row of $H$.
For convenience, $\alpha_i, i\leq d,$ always denotes the architecture parameters after the Softmax transformation.

In our work, the cells themselves serve as building blocks for DARTS as opposed to \cite{liu2018darts} where DARTS is being applied to optimize the architecture of a single cell.
Following \cite{dong2019} and \cite{xie2018} we choose Gumbel-Softmax \cite{jang2016categorical} to incorporate the architecture parameters into the computation path.
Usually, DARTS-based methods consider the architecture parameters as variables over a convex loss surface and optimize them in the same fashion as neural networks weights.
As we want to use the trained architecture parameters as performance predictors, we found it reasonable to firstly regularize the architecture parameters and secondly use a constant learning rate to not interfere with the effect of the former.
Our regularization should serve the purpose of preventing the hyper-architecture from converging too early in favor of certain architectures while supporting such a convergence in later epochs.
To this end, we add the maximum norm on the Gumbel-Softmax regularized architecture parameters to the loss function, scaled by a factor $r\in\mathbb{R}$ linearly decreasing over epochs.
Hence, we update our architecture parameters $\alpha := Gumbel\text{-}Softmax(\alpha^\prime_i)_{i\leq d}, \alpha^\prime\in\mathbb{R}^{d\times w},$ according to the following objective gradient:
\begin{equation}
\label{regularized_loss}
    \nabla_{\alpha^\prime}(L(\alpha, \omega) + r\lVert\alpha\rVert_\infty)
\end{equation}
where $L(\alpha, \omega)$ denotes the neural network loss dependent on architecture parameters $\alpha$ and neural network weights $\omega$.

\textbf{Cell-dependent Regularization}
We elaborate on the regularization factor $r$ in \autoref{regularized_loss} to regularize shallower cells earlier than deeper cells, compare \autoref{fig:cell_regularization} 
This approach is motivated by a cell only being able to learn if its input is somewhat stable.

\ifthenelse{\boolean{requirement_twocolumn}}{
    \begin{figure}
        \centering
        \includegraphics[width=0.9\linewidth]{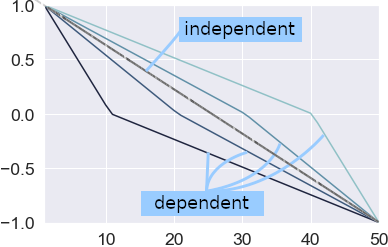}
        \caption{
\textbf{Regularization}
We apply two modes, cell-dependent and cell-independent regularization by means of a regularization factor $r_i\in \mathbb{R}$ (here shown between -1 and 1) along training epochs (here up to 50 as used in the experiments). 
Cell-independent regularization applies a regularization factor $r$ which linearly decreases with increasing epochs (straight linearly decreasing line in the middle).
Cell-dependent regularization, however, applies a differently regularized loss per cell $i$:
$r_i$ decreases faster the smaller $i$.
        }
        \label{fig:cell_regularization}
    \end{figure}
}{}
\ifthenelse{\boolean{requirement_jmlr}}{
}{
    \ifthenelse{\boolean{requirement_twocolumn}}{
        \begin{figure}[tb]
         	\centering
            \includegraphics[width=0.8\linewidth]{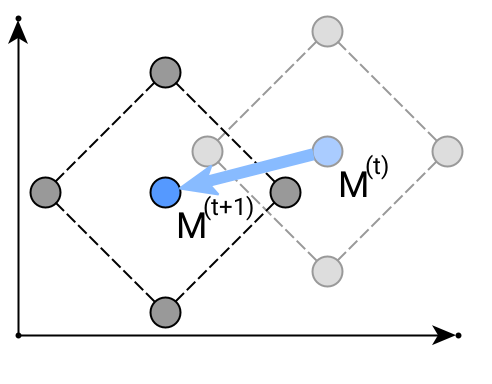}
            \caption{
Depiction of an outer optimization step with gradient descent with finite differences in $\mathcal{S}$ respectively $\mathcal{F}$.
Compare \autoref{eq:anchor-update-finite-differences} for the update of an anchor point $M^{(t)}$ to $M^{(t+1)}$.
Any memory-less search strategy can make use of FaDE-ranks by making a step towards better ranked architectures.
            }
            \label{fig:outer-optimization}
        \end{figure}
    }{
        \begin{figure*}[tbh]
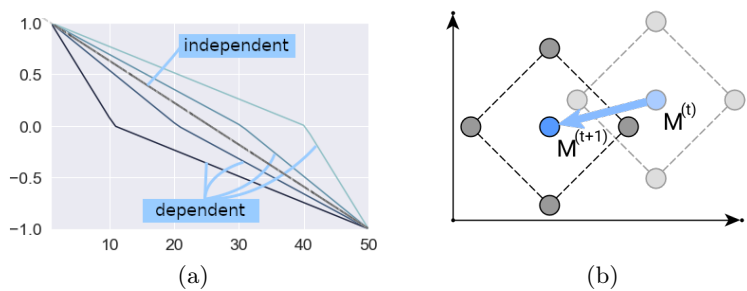

            \vspace{-2em}
            \centering
            \subfloat[]{
                \includegraphics[width=0.4\linewidth]{\visBasePath cell_regularization_combined_annotated.png}
                \label{fig:cell_regularization}
            }
            \qquad
            \subfloat[]{
                \includegraphics[width=0.35\linewidth]{\visBasePath pseudo-gradient-descent-outer-optimization.png}
                \label{fig:outer-optimization}
            }
            \caption{
\small
                \ref{fig:cell_regularization}
\textbf{Regularization}
We apply two modes, cell-dependent and cell-independent regularization by means of a regularization factor $r_i\in \mathbb{R}$ (here in [-1,1]) along training epochs (up to 50). 
Cell-independent regularization applies a regularization factor $r$ which linearly decreases with increasing epochs (linearly decreasing line in the middle).
Cell-dependent regularization, however, applies a differently regularized loss per cell $i$:
$r_i$ decreases faster the smaller $i$.
                \ref{fig:outer-optimization}
Depiction of an outer optimization step with gradient descent with finite differences in $\mathcal{S}$ respectively $\mathcal{F}$.
Compare \autoref{eq:anchor-update-finite-differences} for the update of an anchor point $M^{(t)}$ to $M^{(t+1)}$.
Any memory-less search strategy can make use of \nameFade-ranks by making a step towards better ranked architectures.
            }
            \label{fig:validation}
            \vspace{-1em}
        \end{figure*}
    }
}

\textbf{Weight Sharing}
In addition to implicit weight sharing within the hyper-architecture, implied by a linear number of cells making up for an exponential number of contained architectures, weights might also be shared per row, requiring an implementation of the hyper-architecture where each row itself consists of a hyper-cell that coalesces several cells.
In our experimental setup, for example, where cell architectures are modelled by graphs, we considered the largest graph $G_{max}$ contained in the search space, fully connected, and implemented a hyper-architecture of which each row solely consists of the implementation of $G_{max}$, the hyper-cell.
Now any hyper-architecture can be derived from this hyper-architecture by considering all graphs as sub-graphs of $G_{max}$ and renouncing on devoting strictly separate computation paths to each cell, but instead only virtually separating them as different paths within the hyper-cell.

\subsection{Deriving FaDE-ranks on Hyper-Architecture}
\label{obtaining_fade_ranks}
We use $\alpha$ to rank the subset of architectures contained in $H$ based on
\begin{equation*}
\psi_{\alpha}:H\rightarrow\mathbb{R},(H_{1k_1},\dots,H_{dk_d}) \mapsto \prod_{i\leq d} \alpha_{ik_i}
\end{equation*}
as shown in \autoref{fig:search_space} (middle).
Note that $\psi_{\alpha}$ is just one of many ways to apply the information encoded in $\alpha$ to a ranking of corresponding architectures.
The benefit of this approach arises from the practical - not theoretical - assumption of independence along the depth of an architecture, that allows for predicting an exponential search space in linear time.
Training several architectures $H_{val}\subset H\subset\mathcal{S}^d$ from scratch yields a validation function $\rho$ on $H_{val}$.
A rank correlation coefficient between $\rho$ and $\psi_{\alpha}$, the latter restricted to the validation set, documents how well $\psi_{\alpha}$ predicts relative performances of single architectures contained in $H$.

\subsection{Joint Batch-wise Pseudo Gradient Descent} 
\label{joint_gradient_descent}
A proper correlation between $\rho$ and $\psi_{\alpha}$ indicates the usefulness of the information contained in the $\alpha$ parameters and motivates us to use $\psi_{\alpha}$ in guiding a memory-less, batch-wise search in $\mathcal{S}^d$. 
The downside of the hyper-architecture approach persists in the relative nature, implying that \nameFade-ranks $\psi_{\alpha}$ can in general not be compared among new hyper-architecture evaluations.
Contrary to Pham et al. \cite{pham2018efficient} we argue though that changes in weights of a hyper-architecture eventually invalidate former architecture evaluations within that hyper-architecture and hence any information obtained, whether prediction or real evaluation, inhibits a relative nature anyways, its information content \textit{fading} during search.
Search methods such as gradient descent do not need memory and hence can use the \nameFade-ranks to navigate a NAS.
To apply gradient descent, we assume the search space to be Euclidean and use finite differences on a batch of \nameFade-ranks to approximate a gradient.
Details on the caveat of $\mathcal{S}$ to be Euclidean are being discussed further below.

The overall search works by iteratively sampling hyper-architectures $H^{(t)}, t\in\mathbb{N}$ where the $i$-th row of $H^{(t+1)}$, $i\leq d$ is solely dependent on the $i$-th row of the corresponding architecture parameter $\alpha^{(t)}$, obtained after training $H^{(t)}$.
Compare \autoref{fig:search_space}.
Formally, we consider $d$ independent $w$-dimensional stochastic processes in $\mathcal{S}^w$ and we therefore refer to the search over $\mathcal{S}^d$ being a joint search over $\mathcal{S}$.
The goal is to find hyper-architectures containing well-performing single architectures.
For any row $i\leq d$, an \textit{anchor} point $M^{(t)}_i\in\mathcal{S}$ is being maintained, with $M^{(1)}_i$ being randomly initialized. The cells of row $i$ in $H^{(t+1)}$ are being derived from the standard unit vectors originating from the anchor:
\begin{equation*}
H^{(t)}_i=\{M^{(t)}_i\}\cup\{M^{(t)}_i\pm\gamma*e_k ~|~ k\leq \frac{w}{2}\}
\end{equation*}
where the dimension of $\mathcal{S}$ is assumed to be $\frac{w}{2}$ and where $e_k$ denote the standard unit vectors, $k\leq \frac{w}{2}$.

\begin{wrapfigure}{r}{0.35\linewidth}
    \centering
    \includegraphics[width=0.8\linewidth]{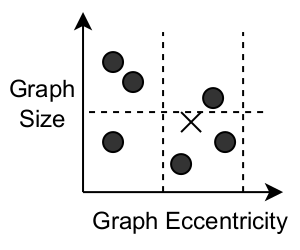}
    \includegraphics[width=0.8\linewidth]{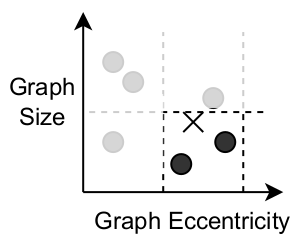}
    \includegraphics[width=0.8\linewidth]{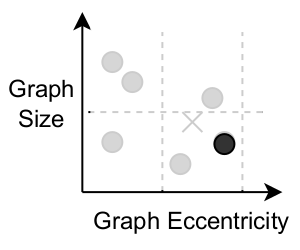}
    \caption{
        \small
        Graph generation:
        a sample in embedding space determines a corresponding bucket from which a graph is drawn.
    }
    \label{fig:feature_space_and_generation}
    \vspace{-5em}
\end{wrapfigure}
The hyper-parameter $\gamma>0$ controls the width of the local environment around the anchor.
After having obtained the architecture parameters for $H^{(t)}_i$ by training $H^{(t)}$, the anchor $M^{(t+1)}_i$ is being derived from descending $M^{(t)}_i$ according to the finite differences along the standard unit vectors:
\begin{equation}
\label{eq:anchor-update-finite-differences}
M^{(t+1)}_i=M^{(t)}_i - \lambda\sum_{k=1}^{\frac{w}{2}}e_k(\beta_i(M^{(t)}_i+\gamma*e_k) - \beta_i(M^{(t)}_i-\gamma*e_k))
\end{equation}
where $\lambda>0$ controls the step size of gradient descent and $\beta_i(\cdot)$ mapping sub-modules to their corresponding architecture parameter after train iteration $i$.

Note that weight sharing could be considered among successive hyper-architectures.
We tested pre-initializing the normal neural network weights of $H^{(k+1)}$ with the trained weights of $H^{(k)}$.

\paragraph{Search Space}
\label{euclidean-space}
We focus on the graph attributes of neural network cell architectures.
Therefore we provide a bijective $embodiment:\mathcal{G}\rightarrow\mathcal{S}$ from a space of directed acyclic graphs to the space of cell architectures.
We let $embodiment$ map a directed acyclic graph to a cell architecture by first prepending an input vertex to vertices with no incoming edges and appending an output vertex to all vertices with no outgoing edges.
The input vertex just serves as interface to distribute the input vector $\boldsymbol{x}$ to all source vertices.
On all edges, except those originating from the input vertex, we place structurally identical convolution layers.
Vertices with more than one input edge combine their inputs by summation and ReLU non-linearity.
While channel count and feature map size within a cell are being fixed, between succeeding cells we approximately double the channel count while proportionally reducing the feature map size.

To enable the pseudo gradient descent from \autoref{joint_gradient_descent}, we use a low-dimensional \textit{feature space} $\mathcal{F}$ and a (stochastic) $generator:\mathcal{F}\rightarrow\mathcal{G}$ that generates graphs from Euclidean vectors.
The generator function is desired to be surjective and smooth with regard to the performance of cell architectures in $\mathcal{S}\cong\mathcal{G}$.
There are several possible choices for $\mathcal{G}, \mathcal{F}$ and $generator$, including $\mathcal{F}$ to be the parameter space of a graph generation algorithm.
In our experiments we settle for simple choices, starting with a rather small $\mathcal{G}$ containing directed acyclic graphs with a low number of vertices.
Hoping on smoothness we consider several scaled graph attributes that have been shown \cite{stier2019structural} to correlate with performance of the corresponding sub-modules in a certain embodiment and training setting, to construct the dimensions of $\mathcal{F}$.

\begin{wrapfigure}{r}{0.35\linewidth}
    \vspace{-2.5em}
    \centering
    \includegraphics[width=\linewidth]{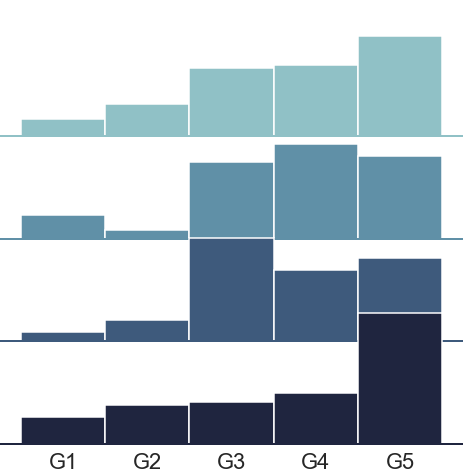}
    \caption{
        Density of softmaxed architecture parameters:
        predicted ranks based on averaged architecture parameter per graph architecture per cell (dark=deep, light=shallow)
    }
    \label{fig:marginals}
    \vspace{-3em}
\end{wrapfigure}

We construct an $embedding: \mathcal{G}\rightarrow\mathcal{F}$ by determining the required graph attributes for each graph of $\mathcal{G}$.
As this embedding is not dense in $\mathcal{F}$, we propose a mapping from the feature space to the space of cell architectures as follows.
$\mathcal{F}$ is being separated into disjunct $dim(\mathcal{F})$-dimensional intervals and define $interval:\mathcal{F}\rightarrow2^\mathcal{F}$ to map a point in the feature space to its containing interval.
We then define $generator$ as $U\circ embedding^{-1}\circ interval$ where $U$ is the discrete uniform sampling.
\autoref{fig:feature_space_and_generation} visualizes the sampling process via $G$ on an exemplary two-dimensional Euclidean feature space.
In addition to potentially increasing sampling speed, dependent on the implementation, the sampling via intervals adds some appreciated noise to the generator.

%
%
\paragraph{Related Work}
\label{sec:related-work}
\cite{cai2018} extend DARTS \cite{liu2018darts} by sampling a single embedded architecture for each forward pass instead of taking a weighted sum which, besides reducing memory, might even result in more reliable training as the sparsity of a forward pass is the same for hyper-architecture and target architecture.
Also \cite{dong2019} and \cite{xie2018} feature a sparse forward pass by using Gumbel-Softmax \cite{jang2016categorical} instead of a weighted sum for aggregating parallel architecture choices.
\cite{Chen2019} progressively drop the weakest connections as training progresses to reduce memory footprint.
They use the saved resources to progressively increase depth of the hyper-architecture w.r.t. stacked sub-modules and thus aim at closing the gap between proxy and target domain.
However, they feature only copies of the same sub-module.
\cite{hao2021layered} propose a progressive approach that successively searches a stack of different sub-modules.
\textbf{Progressive} approaches are also used without DARTS, e.g. by starting from a space containing very small network modules and end up in a space containing modules of desirable size \cite{liu2018prog}.
This restricts search space exploration to a constant complexity in each iteration by step-wise building up from well performing smaller modules.
\textbf{Weight sharing} comes implicit in DARTS but is also used in other work, i.e. \cite{pham2018efficient} equip a recurrent NAS pipeline with a hyper-architecture such that instead of training discrete architectures proposed by the pipeline from scratch, they are initialized with the corresponding weights of the hyper-architecture.
That enhancement is applicable out of the box to most NAS approaches and shows a significant decrease in resource usage while achieving comparable results.

%
%
\section{Experiments}
\label{sec:experiments}
We consider a multi-cell search space consisting of $n_c=4$ chained cells, each cell featuring a DAG with less than $n_v=6$ nodes as cell architecture.
In a first experiment we obtain \nameFade-ranks on a single hyper-architecture and show that they correlate well with the actual performances of a small subset of architectures contained in the hyper-architecture. 
Another experiment iteratively trains hyper-architectures according to \autoref{joint_gradient_descent} in order to search the complete search space.
For the latter we present that search results improve over iterations, though not significantly.

\begin{wrapfigure}{l}{0.4\linewidth}
    \vspace{-1em}
    \centering
    \includegraphics[width=\linewidth]{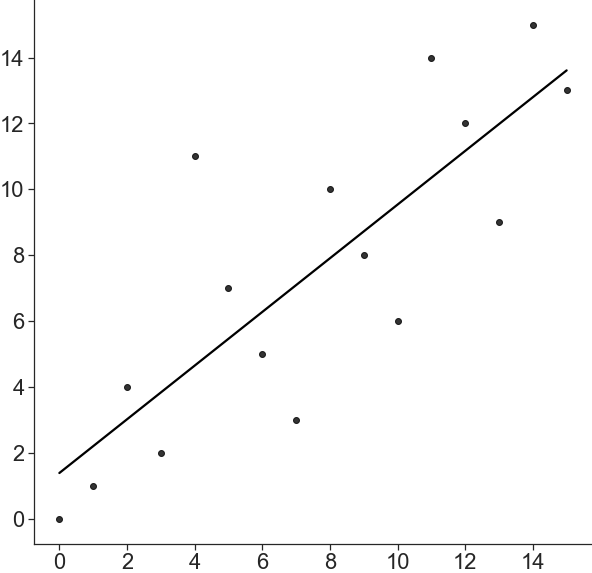}
    \caption{
        Correlation between predicted and evaluated ranks.
    }
    \label{fig:results_b_spearman_cof_c18i}
    \vspace{-1.5em}
\end{wrapfigure}

\noindent\textbf{Dataset Preprocessing}
All experiments are conducted on the CIFAR-10 image classification dataset.
The employed dataset split is $1-1-4$, meaning one part was used for \emph{testing}, one part for \emph{architecture training} and four parts for \emph{weight training}.
Between succeeding cells, max pooling is applied for downsampling.
All convolutional cells have been using the same kernel shape of $5\times 5$.
Gradient clipping during backpropagation was employed with an absolute value of $10$.

We used the following \textbf{hyperparameters} across all experiments and provide a \href{\urlNotebook}{github repository} for detailled reproducibility:
activation function is ReLU, loss is cross-entropy, DARTS aggregation is hard Gumbel-Softmax, the aggregation func. temperature is ten.
We used 16 or 32 channels for deepest cells and Max() as pooling operation.
For optimization we used a batch size of 128, Adam with $\beta_1=0.1, \beta_2=10^{-3}$, $\alpha=10^{-3}, \epsilon=10^{-8}$, a weight decay of $10^{-4}$, Kaiming as weight initialization, $\mathcal{N}(0; 0.5)$ for the architecture initialization and a gradient clipping value of ten.

\subsection{Validating FaDE-Ranks \jupyterLink{\urlNotebookA}}
\label{validating_fade}
We construct a hyper-architecture that with $n_g=5$ manually selected DAGs as parallel computation paths per cell.
Once embedded, each DAG comes with the same number of weights, approximately $0.25\cdot 10^6$.
The hyper-architecture $H$ spans a finite search space with $n_g^{n_c}=5^4=625$ architectures.
After training the $H$, we obtain \nameFade-ranks as described in \autoref{obtaining_fade_ranks}.

We consider the architecture parameter distributions per cell, averaged over several experiment repetitions, as independent marginals of a joint discrete distribution on the finite architecture search space.
\autoref{fig:marginals} illustrates the predicted marginal distributions of the trained hyper-architecture.

Training a subset of $16$ manually selected discrete architectures enables calculating a spearman rank correlation between \nameFade-ranks and evaluation ranks, as shown in \autoref{fig:results_b_spearman_cof_c18i}.
The correlation coefficient of $~0.8$ is significant and shows that our methodology predicts the relative performances within the small architecture subset quite well.
That means, the obtained \nameFade-ranks can be used to guide an open-ended search with local information.

\begin{wrapfigure}{r}{0.4\linewidth}
    \vspace{-1em}
    \centering
    \includegraphics[width=\linewidth]{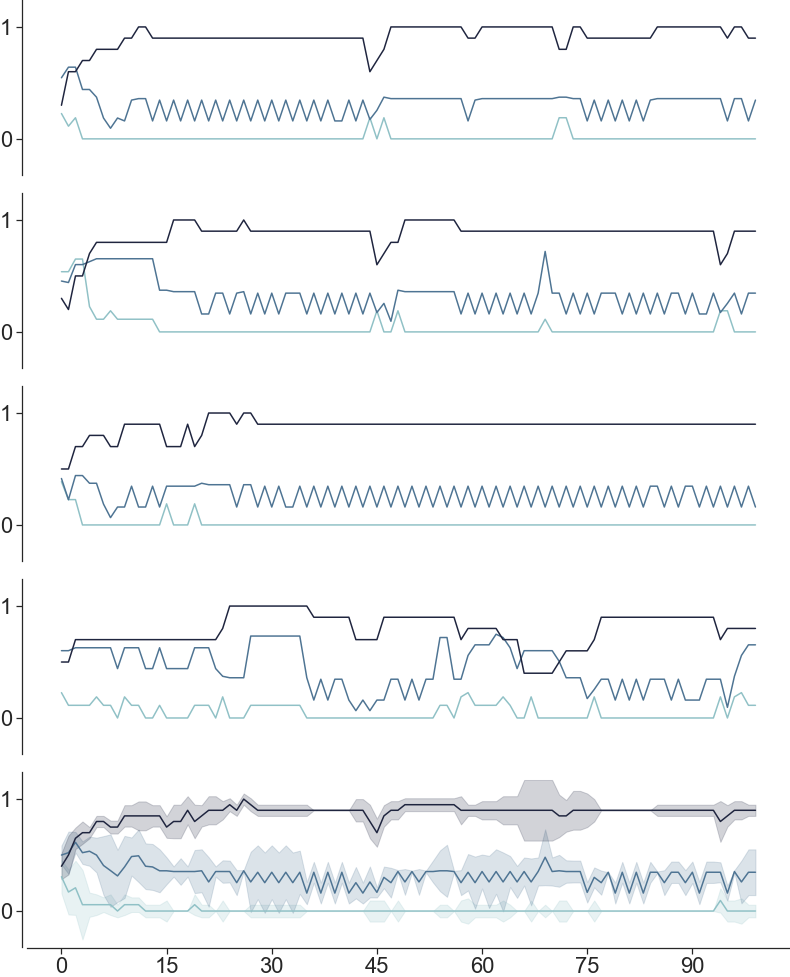}
    \caption{
        \small
        Search trajectories in $\mathbb{R}^3$ per cell per dimension:
        per cell the anchor point of the pseudo gradient descent is plotted for $100$ epochs.
        The anchor point coordinates are color-coded by feature space dimension which share a common y-axis as their values are normed to the same interval.
        We observe a rough convergence of trajectories.
    }
    \label{fig:outer_trajectories}
    \vspace{-5em}
\end{wrapfigure}

From further experiments we observed that weight sharing decreases the obtained correlation compared to the results in \autoref{fig:results_b_spearman_cof_c18i}.
However, weight sharing yields quite significant correlation results already with a few number of training epochs.
This may be owed to the fact, that weight sharing implies weights to be trained more often and thus counteracts the reduction in epochs.
We will resort to a fewer number of training epochs while sharing weights in later experiments.

\subsection{NAS on iterative FaDE-Ranks \jupyterLink{\urlNotebookB}}
\label{sec:nas-iterative}
\begin{figure}[tb]
    \vspace{-1em}
    \centering
    \includegraphics[width=0.9\linewidth]{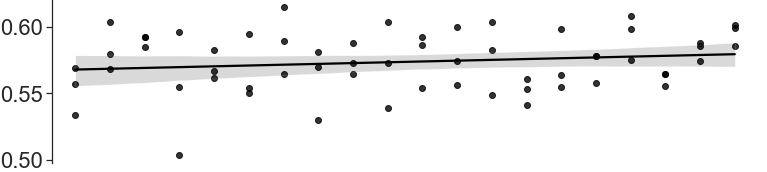}
    \caption{
        \small
        Evaluation accuracy of architectures generated from points in the search trajectory:
        The evaluation performance is slightly increasing
    }
    \label{fig:outer_regression}
    \vspace{-1.5em}
\end{figure}

We aim at iteratively improving the cell architectures of the hyper-architecture from \autoref{validating_fade} according to the methodology described in \autoref{joint_gradient_descent}.
A pseudo gradient descent serves as optimization strategy on the feature space $\mathcal{F}$.
We consider architectures with $4$ cells, assign one feature space per cell and run independent, parallel pseudo gradient descent algorithms on them.
The \textit{joint feature space} refers to the product of the $4$ feature spaces.
Building on \cite{stier2019structural}, $e:\mathcal{G}\rightarrow\mathcal{F}$ maps a DAG to three dimensions according to its normed eccentricity variance, degree variance and number of vertices, such that $\mathcal{F}=\mathbb{R}^3$.

A single experiments runs with 100 epochs, each featuring five epochs of hyper-architecture optimization.
For each cell $i=1,\dots,4$, we obtain feature space trajectories $(M^{(t)}_i)_{t=1,\dots,100}$.
In order to validate these trajectories, in regular intervals of $t$, we repeatedly construct an architecture from the DAGs generated from $M^{(t)}_i, i=1,\dots,4,$ and evaluate it using $30$ epochs of architecture training.
We thus obtain a validation function from the trajectory index space into $\mathbb{R}$.
An increasing function, coarsely measurable by a linear regression on its index, would indicate our search strategy to yield better architectures over time.
\autoref{fig:outer_regression} shows individual model evaluations, including a linear regression.
The pearson coefficient of $0.16$ is barely significant.

\autoref{fig:outer_trajectories} shows the trajectories in all three dimensions per cell from shallow to deep and its bottom graphic shows the trajectories aggregated over cells.
We interestingly notice that the trajectories already show convergence within $20$ epochs.
Note that convergence in \textit{number of operations} towards the high end of the scale occurs in every cell.
This convergence is in accordance with what one would naturally expect.
We also observe that the deeper the cell the more divergent its trajectories.
There are artifacts that could be attributed to the sparsity of our search space or poor heuristics of our search space sampling, for example the large step sizes of the trajectories and its occasional chainsaw pattern.

\begin{figure}[tb]
    \vspace{-1em}
    \centering
    \includegraphics[width=0.9\linewidth]{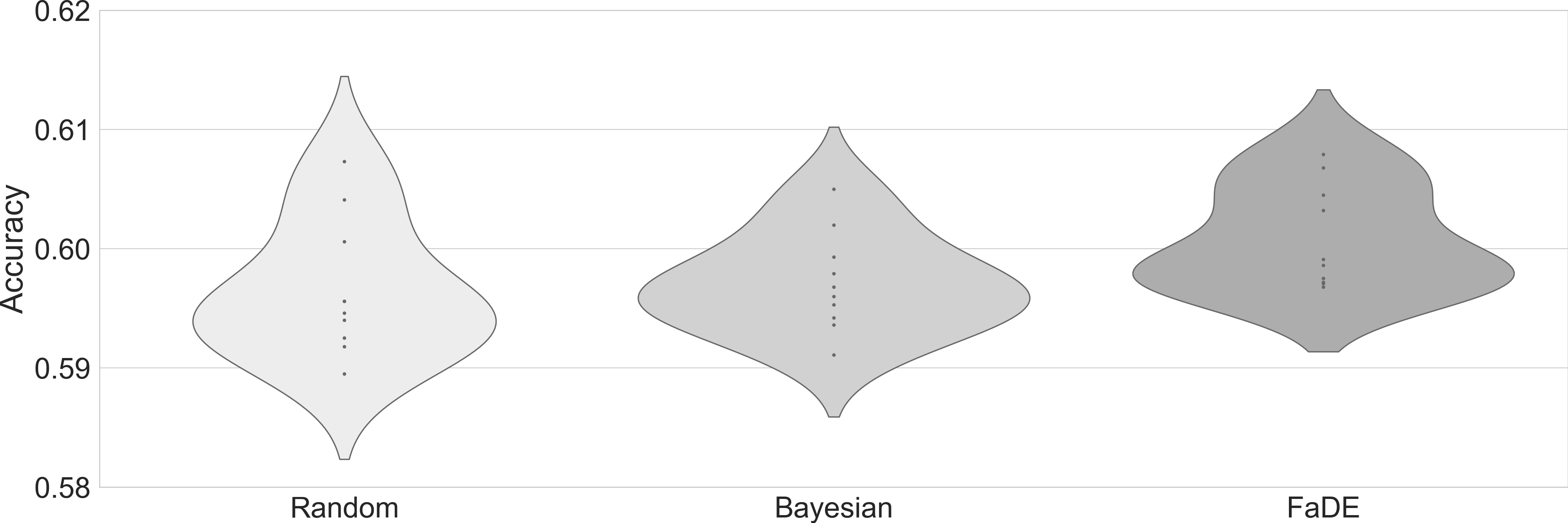}
    \caption{
        \small
        Comparing the distribution of top architectures found by random search (\textbf{RS}), bayesian optimization (\textbf{BO}) and \nameFade on $\mathcal{F}$ spanned by variances of eccentricity, degree and number of vertices.
        \nameFade finds more well-performing architectures in less time.
        RS provides a baseline which clearly shows a higher standard deviation across found accuracy scores.
        BO is a common NAS method, especially when searching through a low-dimensional search space as we used it.
    }
    \label{fig:benchmark}
    \vspace{-1.5em}
\end{figure}

To validate the outer Neural Architecture Search, we compared to random search (\textbf{RS}) and bayesian optimization (\textbf{BO}).
For 50 epochs, the search methods propose a point in the joint $\mathcal{F}$ for evaluation.
The median accuracy of five such architecture generations and evaluations is being fed back to the search algorithm.
\textbf{BO}, similar to the pseudo gradient descent on \nameFade, assumes its stochastic models per cell to be independent from each other.
Even though, this is not the case, we argue that the number of epochs is too small for a more complex bayesian model that does not make this assumption of independence.
For \textbf{BO}, we use the \textit{Upper Confidence Bounds} method with $\kappa=2.5$, $\xi=0$.
We compare the results with the pseudo gradient descent validation results, this time validating the trajectories for the first $25$ of $100$ outer epochs.
We do not use $50$ epochs for the pseudo gradient descent on \nameFade as the trajectories already converge earlier.
\autoref{fig:benchmark} provides test accuracies of the top $10$ epochs per search method.

Points in \autoref{fig:benchmark} are a median of five accuracy evaluations, generated from the same point in $\mathcal{F}$.
The distribution of accuracy evaluations of a single point represents an important criteria on the suitability of the search space.
A wide-spread distribution indicates that the feature space does not capture well architecture features that correlate with the corresponding evaluation performance. 
The mean standard deviation across all data points in our plot is $0.014$ which is quite high compared to the observed magnitudes in \autoref{fig:benchmark}.


%
%
\section{Conclusion \& Future Work}
\label{sec:conclusion}
We presented \nameFade, a method to leverage differentiable architecture search to aggregate path decisions from a fixed hierarchical hyper-architecture into point estimates for an open-ended search.
The aggregated estimates are called \nameFade-ranks and show a positive rank correlation with individually trained architectures.
Justified with this correlation, \nameFade-ranks can be used to guide an outer search in a pseudo-gradient descent manner.
The method is generalizable in a way that alternative strategies for the outer search can be employed as long as the relative nature of the ranks are respected.
We see future work in both \textbf{1/} the analysis of the quality rank information for global search as well as \textbf{2/} experiments with more complex graph feature spaces, e.g. obtained from generative graph models.

\bibliographystyle{splncs04}

\end{document}